\documentclass{llncs}

\usepackage{graphicx} 
\usepackage[T1]{fontenc}
\usepackage{amsmath}
\usepackage[english]{babel}
\usepackage{array,tabularx}
\usepackage{booktabs} 
\usepackage{makeidx}  
\usepackage{xparse}
\usepackage{moredefs}
\usepackage{lips}
\usepackage{epstopdf} 
\usepackage{color}
\usepackage{csquotes}
\usepackage{xcolor}
\usepackage{times}
\usepackage[hidelinks]{hyperref}
\usepackage[nolist,nohyperlinks]{acronym}
\usepackage{comment}
\usepackage{paralist}
\usepackage{cleveref}
\usepackage{wrapfig}
\usepackage{multirow}
\usepackage{microtype}
\usepackage{listings}
\usepackage[super]{nth}
\usepackage{natbib}
\usepackage{pgfplots}
\usepackage{pgfplotstable}
\pgfplotsset{compat=1.18, every axis label/.append style={font=\scriptsize}}
\usepackage{subcaption}
\usepackage{graphicx}
\usepackage{dirtytalk}

\usepackage{float}
\floatstyle{plaintop}
\restylefloat{table}

\usepackage[style=base,labelfont=bf,labelsep=period,tableposition=top,font=small]{caption}
\makeatletter
\renewcommand\@biblabel[1]{#1.}
\makeatother
\addto\captionsenglish{}

\usepackage{fancyhdr}
\fancypagestyle{WI_footer}{\fancyhf{}\fancyfoot[L]{\color{black}\scriptsize submitted to 20th International Conference on Wirtschaftsinformatik\\September 2025, Münster, Germany}}

\crefformat{footnote}{#2\footnotemark[#1]#3}
\expandafter\def\expandafter\UrlBreaks\expandafter{\UrlBreaks
  \do\a\do\b\do\c\do\d\do\e\do\f\do\g\do\h\do\i\do\j%
  \do\k\do\l\do\m\do\n\do\o\do\p\do\q\do\r\do\s\do\t%
  \do\u\do\v\do\w\do\x\do\y\do\z\do\A\do\B\do\C\do\D%
  \do\E\do\F\do\G\do\H\do\I\do\J\do\K\do\L\do\M\do\N%
  \do\O\do\P\do\Q\do\R\do\S\do\T\do\U\do\V\do\W\do\X%
  \do\Y\do\Z}

\newcolumntype{L}[1]{>{\raggedright\arraybackslash}p{#1}}   
\newcolumntype{C}[1]{>{\centering\arraybackslash}p{#1}}     
\newcolumntype{R}[1]{>{\raggedleft\arraybackslash}p{#1}}    

\begin{document}
\frontmatter          

\mainmatter              

\title{Evaluating the Process Modeling Abilities of Large Language Models -- Preliminary Foundations and Results}
\subtitle{Research in Progress} 

\author{Peter Fettke\inst{} \and
Constantin Houy\inst{}}

\institute{German Research Center for Artificial Intelligence (DFKI) and\\ Saarland University, Saarbrücken, Germany\\
\email{\{peter.fettke,constantin.houy\}@dfki.de}}

\maketitle
\setcounter{footnote}{0}

\begin{abstract}
Large language models (LLM) have revolutionized the processing of natural language. Although first benchmarks of the process modeling abilities of LLM are promising, it is currently under debate to what extent an LLM can generate good process models. In this contribution, we argue that the evaluation of the process modeling abilities of LLM is far from being trivial. Hence, available evaluation results must be taken carefully. For example, even in a simple scenario, not only the quality of a model should be taken into account, but also the costs and time needed for generation. Thus, an LLM does not generate one optimal solution, but a set of Pareto-optimal variants. Moreover, there are several further challenges which have to be taken into account, e.g. conceptualization of quality, validation of results, generalizability, and data leakage. We discuss these challenges in detail and discuss future experiments to tackle these challenges scientifically.\\

\end{abstract}

\thispagestyle{WI_footer}



\section{Motivation}

Large language models (LLM) have revolutionized many tasks of natural language processing (NLP). Recently, LLM are not only used for language related tasks, but also for generating program code \citep{joel2024surveyllmbasedcodegeneration, sarkar2022likeprogramartificialintelligence}, planning \citep{katz2024thought, erdogan2025planandactimprovingplanningagents}, action models \citep{wang2025largeactionmodelsinception, zhang2024xlamfamilylargeaction}, and many more.

In 2023, first ideas to use LLM for generating and understanding conceptual models emerged \citep{Fill2023Conceptual}. Since then, different approaches for the use of LLM have been developed and tested. The latest developments aim at using LLM for interactive process modeling, e.g. \citep{10.1007/978-3-031-61007-3_18}, for process discovery from event logs \citep{norouzifar2024bridging}, for improving process model understandability \citep{kourani2024leveraginglargelanguagemodels}, and other tasks related to process modeling \citep{DBLP:conf/bpm/VidgofBM23}.

Although the machine-generated results are often surprising and of astonishing quality, it is obvious that the evaluation has to be conducted in some objective way. For such a systematic evaluation, already first benchmark studies have been undertaken and published \citep{kourani2024evaluating}.

However, in this article, we argue that an objective evaluation is far from trivial for a number of reasons, e.g. model quality is multi-criteria measure \citep{10.1007/978-3-031-77908-4_7, books/sp/Krogstie16, 10.1007/978-3-319-19237-6_25} with dimensions such as syntactic correctness, semantic adequacy, and understandability \citep{10.1007/978-3-642-34002-4_5, 10.1007/978-3-030-35646-0_11}. The long-term objective of this piece of research in progress aims at developing the necessary foundations and methods for an objective evaluation of the process modeling abilities of LLM. As a first step to tackle the evaluation challenges, we define a typical standard evaluation scenario. Based on this standard scenario, we are able to elaborate on particular evaluation challenges and possible ways to overcome them. 

In the following, we report some preliminary results, namely, the assumptions of the standard evaluation scenario (section 2); the conceptualization of quality measures describing the LLM performance and the trade-off between quality, cost, and time (section 3); a specific evaluation example (section 4); the discussion of related work (section 5); and conclusions and discussion of future work (section 6).

\section{Assumptions of the standard evaluation scenario}

We introduce three assumptions to define the standard evaluation scenario:

\begin{enumerate}

\item A process model conceptualizes a real or imagined modeling domain. We assume that the domain to be modeled is given by a plain English text. In other words, the task to be accomplished by the LLM is to transform or reformulate the natural textual description by using a process modeling language. 

\item For each textual domain description one or more sample solutions are given. They serve as a basis for model evaluation and can be considered as the \say{gold-standard}; in machine learning, the gold-standard is typically called \say{ground-truth}.

\item The sample solutions of the gold-standard use the Business Process Modeling Notation (BPMN) as a process modeling language.
\end{enumerate}

It is obvious that in many real-world modeling situations the three assumptions are not fulfilled. However, many arguments justify them, e.g. BPMN is often treated as the de-facto standard for process modeling \citep{VONROSING2015433}, a textual domain description provides a controlled environment which is necessary for the comparison of results, numerous examples demonstrate that the generation of a process model based on a textual description is realistic to a certain degree and feasible for humans and machines. However, later we will discuss some consequences for the evaluation if one or more of these assumptions are relaxed in particular ways.

\section{Measuring the performance in the standard evaluation scenario}

\subsection{Quality}

Results from the field of NLP clearly demonstrate that measuring the quality of a machine-generated language artifact and comparing it with the quality of a human-generated task is far from simple, e.g. evaluating the subtle language differences in machine translation. We argue that the challenges of evaluation in the standard process modeling scenario are comparable to the challenges known in the field of NLP.

One simple solution, also often used in NLP research, relies on the opinions of domain or modeling experts. Such experts are often acquired using well-known crowd-working platforms. Such an approach is inherently subjective and it is not clear how to control the quality of the experts' work, particular in large scale evaluations.

More objective measures are also problematic. While syntactic deficits can relatively easily be counted, semantic differences are much more difficult to evaluate, e.g. different abstraction levels used for modeling, focusing on relevant aspects or the omission of irrelevant details. Last but not least, it could be argued that pragmatic aspects of the model must also be taken into account, e.g. the layout of a BPMN diagram, the intended purpose of the model or the task the modeller would like to solve with the model.

Similar to typical measures used in information retrieval, e.g. precison and recall, the quality of the model in question could be measured: How many concepts of the model are not contained in the gold-standard? How many concepts of the gold-standard are not in the model?

The previous idea can be conceptualized more concrete, if an execution model of a BPMN diagram is additionally assumed: It could be argued that all execution paths of the gold-standard should be included in the presented solution \citep{kourani2024evaluating}. On the other hand, execution paths of a particular solution which are not possible in the gold-standard should be penalized. Note, that such an approach relies on the use of an appropriate matching between the gold-standard and the provided solution -- and process matching is far from easy \citep{DBLP:conf/mkwi/ThalerHFL14}.

Since BPMN diagrams can be interpreted as graphs, the idea of a graph-edit-distance \citep{bunke1998graph} can be employed \citep{DIJKMAN2011498}: How many edit operations are necessary to transform the provided solution into the gold-standard model? This idea is quite simple and can theoretically be easily interpreted: the quality of a machine-generated solution is correlated to the amount of rework needed to revise the model accordingly. The particular needed edit operations could also be weighted based on a specific cost model for each operation.

Currently, although the described ideas for measuring the quality has some face validity, the validity of these measures for evaluating process modeling abilitites of LLM is more or less unknown.

\subsection{Time and costs}

In addition to the quality of the generated solution, the time for generating a model is of importance. For such analyses, theoretical computer science developed a rich theory supporting the analysis of algorithms based on different inputs; the so-called \say{Big O notation}. However, currently we do not know if such a theoretical analysis produces valid results in our case and we are not aware of any work which is relevant in this context. As a simple alternative, the time an LLM needed for generating the output model can easily be measured in an evaluation procedure.

Furthermore, the usage of LLM is not without costs. First, there are some fix costs for setting up the infrastructure and training the LLM. Second, there are variable costs for using LLM. One major idea of LLM are so-called foundational models which can be used for different tasks and are offered as services by different technology companies \citep{bommasani2022opportunitiesrisksfoundationmodels,Schneider2024FoundationModels}. The variable costs for usage are typically measured in input tokens, output tokens, API calls, etc., and have to be paid in US-Dollars, Euros, or other currencies. Hence, it is also possible to quantify the costs of using an LLM in such service scenarios. However, note, that the particular variable costs of using foundational models in evaluating the abilities of LLM should not be neglected; e.g. \cite{stein2025automatinggenerationpromptsllmbased} do not benchmark every combination of an experimental study design for evaluating the planning abilities of LLM because the estimated costs exceeded their budget. This is not without practical relevance, particular in comparison for the labor costs needed for a human-generated model.

One further aspect is of importance: Since LLM do not work deterministically, a repetitive measurement of all measures is needed and some averaging is necessary. Therefore, systematic studies of LLM parameters such as the so-called \say{seed} and \say{temperature} of the model must be studied in a systematic way.

\section{An example}

\begin{figure}[ht]
    \centering
    \includegraphics[width=1.0\textwidth]{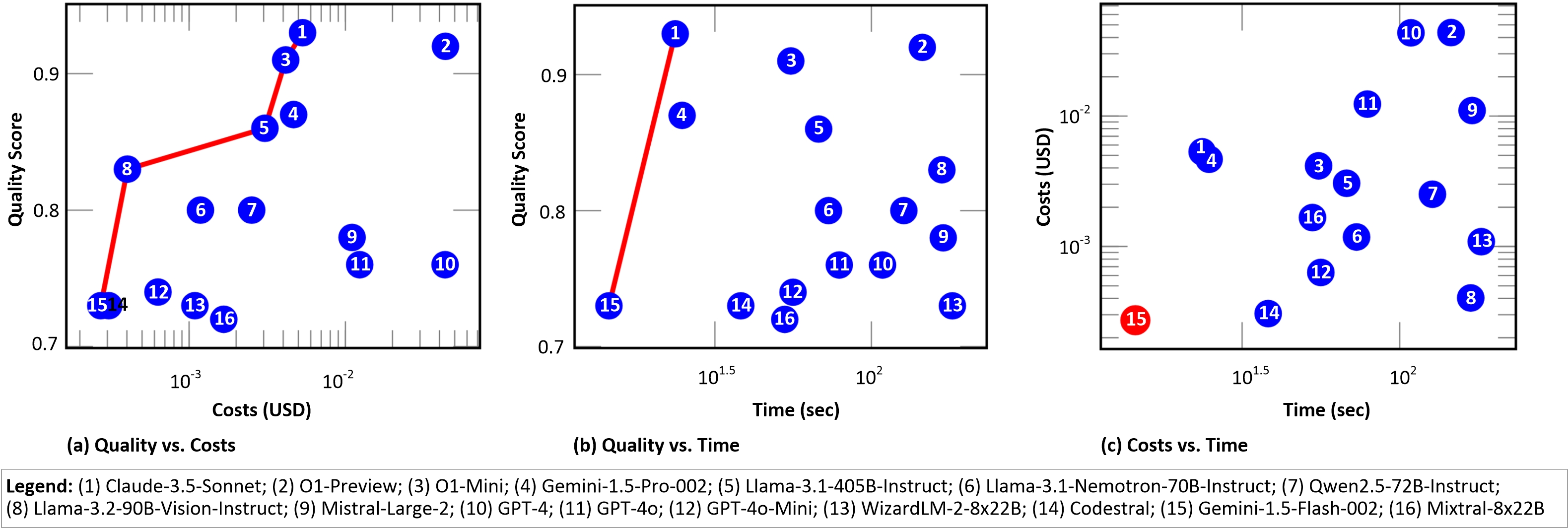}
    \caption{Pareto analysis for Quality, Cost, and Time (Pareto fronts in red, log-scaled axes)}
    \label{fig:1}
\end{figure}

So far, we have not established the full infrastructure for presenting a completed benchmarking study. Hence, the presentation of the results of a full automation of the process is currently not possible. However, we aim at a fully-automated pipeline for the above-mentioned standard case. Previous work on automated assessment of modeling results demonstrates that automatically measuring quality is possible in principle, e.g. \cite{thaler2016automated}, \cite{ullrich2023automated}.
In addition to that, the cost to compute the benchmarks plays an important role. However, the development of approaches is progressing rapidly and many different fine-tuned LLM supporting different specific tasks exist. Against this background, it is currently questionable how science can keep up to produce a truly acceptable assessment of approaches. In turn, this does not mean that we should not try. Documented experience with several LLM shows that for some textual descriptions, acceptable or even good results were produced. Inspired by the study of \cite{kourani2024evaluating}, we used their results on the quality and time efficiency of \say{a diverse set of state-of-the-art LLM} for process modeling tasks [p.11] and extended them by adding the cost dimension measured in US-Dollars. In this context, we let GPT-4 estimate the token consumption of each LLM based on the given average time elapsed for producing an acceptable model in the above study, checked them for plausibility, and calculated the resulting costs (in US-Dollars) using an extensive and commonly-used LLM pricing calculator (https://huggingface.co/spaces/Presidentlin/llm-pricing-calculator) based on the prices of providers of each LLM. The following diagrams visualize our results in combination with the results of \cite{kourani2024evaluating} and present three Pareto fronts for the perspectives \say{quality vs. cost}, \say{quality vs. time} and \say{cost vs. time}. In the latter, there is one optimal data point concerning cost vs. time, hence the Pareto front is degenerated and there is only one Pareto-optimal solution in this example (see Figure \ref{fig:1}).

\section{Related work}

We relate our work to five research streams, namely (1) automated modeling, (2) automated model assessment, (3) evaluation of abilities of LLM in general, (4) known evaluations of process modeling abilities of LLM, and (5) available data sets.

(1) While early work on automated modeling used rule-based approaches, recent articles investigate the potential of LLM for generating process models from the content of documents \citep{voelter2024leveraginggenerativeaiextracting} or using other formats of process description, like declarative process descriptions which can be automatically transformed into BPMN models \citep{10.1007/978-3-030-11641-5_40}.
(2) An overview of works on the automated assessment of conceptual models as well as process models in education can be found in \cite{ullrich2023automated}. For example, \cite{thaler2016automated} present an approach for the automated assessment of process model quality with regard to syntax, semantics, and pragmatic quality, e.g. understandability, for the example of Event-driven Process Chains (EPC). \cite{Westergaard2013180} present an according approach for coloured Petri nets, \cite{Sanchez-Ferreres2020552} for BPMN, and \cite{Schramm2012472, Striewe2014336} and \cite{Beck2015} focus on UML activity diagrams in the context of teaching process modeling.
(3) The evaluation of abilities of LLM in general is treated in many different works proposing different approaches for different domains, e.g. \cite{hu2024unveilingllmevaluationfocused, 10.1115/1.4066730}. Furthermore, evaluation foundations are generally discussed, e.g. by \cite{10.1162/tacl_a_00681} and \cite{kapoor2024ai}.
(4) The so far most pertinent work in the context of studying the evaluation of LLM in process modeling is the study by \cite{kourani2024evaluating} which has taken a look at quality and time, but not at costs; they also do not study the trade-off between relevant measures or identify a Pareto front. Therefore, we complemented this study with our additional perspective on modeling costs and there relationship.
(5) Another important line of work is on data which can be used as benchmarks for modeling, e.g. \cite{DBLP:conf/ejc/WalterTAFL14,DBLP:conf/bpm/BellanADGP22,DBLP:conf/icpm/SolaWSBRK22}. Not all of these datasets do contain a textual description of the domain to be modeled but could be added with generative synthetic approaches.

\section{Conclusions and further research work}

We presented the design for an experimental laboratory study of the process modeling abilities of LLM. Compared to a field study, always validity questions arise, e.g. regarding the representativeness of the standard modeling scenario. Although the used modeling scenarios are divers, we do not have a precise qualification for the validity of the samples. Since known evaluations only use more or less convenient samples, the results should be treated with caution.

In fact, it can be argued that the LLM and training datasets used are inherently problematic. Since many available foundation models are not open and transparent (\cite{Liesenfeld_2023,huang2024trustllmtrustworthinesslargelanguage}), there is a chance that the foundational models used the known datasets for their own training. Formulated in more technical terms, if the well-known datasets are used for evaluation, the percentage of data leakage might be 100 percent. Without transparency, the comparison and evaluation of the abilities of LLM is almost uncontrollable and thus biased to an unknown extent. This problem is well-known in the context of the domain of protein prediction \citep{marx2022method}. For an adequate evaluation of a system for protein prediction, biomolecules with unknown protein structures must be used. Otherwise, it cannot be excluded that the LLM is \say{cheating} because the correct solution is known and contained in the training data set. Hence, systematic studies of the generalization abilities of LLM are needed.

Furthermore, the introduced assumptions of the standard evaluation scenario can be relaxed. First, BPMN has a primary focus on the flow of activities and does not explicitly model data objects. However, it has been well known for decades that data is of major importance to understand business processes \citep{DBLP:journals/is/Scheer94,DBLP:books/sp/Weske19}. Hence, other modeling languages might be used to evaluate the results of an LLM.

Secondly, instead of a textual description, other modalities for providing information about the domain of modeling can be provided, e.g. images, video, or speech \citep{CHVIROVA2024110716}. In this direction, it might be interesting to set up a virtual laboratory environment to detect and model business processes, e.g. \cite{DBLP:conf/bpm/KnochPS20} already set up such scenarios, but do not use them for the evaluation of LLM modeling abilities.

Third, it might be interesting not only to use the abilities of an LLM for solving the standard modeling scenario. Instead process modeling may be assisted by different LLM, e.g. evaluating the solution, providing feedback, gaining data, improving the solution, translating natural language used for domain description, or interactively training the LLM. It is obvious that such an interactive modeling scenario (\say{agentic process modeling}) is attractive but cannot easily be controlled in real-world or laboratory scenarios.

Fourth, the previous aspect already focuses on particular application domains of process modeling. Since the vocabulary and the workflows in process modeling are strongly dependent on the application domain, it might be necessary to focus on particular domains for evaluation, e.g. manufacturing, health, insurance. If a particular domain is focused, it might be necessary to fine-tune the LLM for this particular purpose.

In summary, a valid evaluation of the abilities of an LLM for process modeling is not only of practical relevance but naturally leads to several theoretically interesting questions for the field of process modeling.



\bibliographystyle{agsm}
\bibliography{main}

@misc{stein2025automatinggenerationpromptsllmbased,
      title={Automating the Generation of Prompts for LLM-based Action Choice in PDDL Planning}, 
      author={Katharina Stein and Daniel Fišer and Jörg Hoffmann and Alexander Koller},
      year={2025},
      journal   = {arXiv preprint arXiv:2311.09830} 
}

@inproceedings{Liesenfeld_2023, series={CUI ’23},
   title={Opening up ChatGPT: Tracking openness, transparency, and accountability in instruction-tuned text generators},   
   DOI={10.1145/3571884.3604316},
   booktitle={Proceedings of the 5th International Conference on Conversational User Interfaces},
   publisher={ACM},
   author={Liesenfeld, Andreas and Lopez, Alianda and Dingemanse, Mark},
   year={2023},
   month=jul, pages={1–6},
   collection={CUI ’23} }

@article{huang2024trustllmtrustworthinesslargelanguage,
      title={TrustLLM: Trustworthiness in Large Language Models}, 
      author={Yue Huang and Lichao Sun and Haoran Wang and Siyuan Wu and Qihui Zhang and Yuan Li and Chujie Gao and Yixin Huang and Wenhan Lyu and Yixuan Zhang and Xiner Li and Zhengliang Liu and Yixin Liu and Yijue Wang and Zhikun Zhang and Bertie Vidgen and Bhavya Kailkhura and Caiming Xiong and Chaowei Xiao and Chunyuan Li and Eric Xing and Furong Huang and Hao Liu and Heng Ji and Hongyi Wang and Huan Zhang and Huaxiu Yao and Manolis Kellis and Marinka Zitnik and Meng Jiang and Mohit Bansal and James Zou and Jian Pei and Jian Liu and Jianfeng Gao and Jiawei Han and Jieyu Zhao and Jiliang Tang and Jindong Wang and Joaquin Vanschoren and John Mitchell and Kai Shu and Kaidi Xu and Kai-Wei Chang and Lifang He and Lifu Huang and Michael Backes and Neil Zhenqiang Gong and Philip S. Yu and Pin-Yu Chen and Quanquan Gu and Ran Xu and Rex Ying and Shuiwang Ji and Suman Jana and Tianlong Chen and Tianming Liu and Tianyi Zhou and William Wang and Xiang Li and Xiangliang Zhang and Xiao Wang and Xing Xie and Xun Chen and Xuyu Wang and Yan Liu and Yanfang Ye and Yinzhi Cao and Yong Chen and Yue Zhao},
      year={2024},
      journal   = {arXiv preprint arXiv:2401.05561} 
}

@inproceedings{DBLP:conf/bpm/VidgofBM23,
  author       = {Maxim Vidgof and
                  Stefan Bachhofner and
                  Jan Mendling},
  editor       = {Chiara Di Francescomarino and
                  Andrea Burattin and
                  Christian Janiesch and
                  Shazia Sadiq},
  title        = {Large Language Models for Business Process Management: Opportunities
                  and Challenges},
  booktitle    = {Business Process Management Forum - {BPM} 2023 Forum, Utrecht, The
                  Netherlands, September 11-15, 2023, Proceedings},
  series       = {Lecture Notes in Business Information Processing},
  volume       = {490},
  pages        = {107--123},
  publisher    = {Springer},
  year         = {2023},
  doi          = {10.1007/978-3-031-41623-1\_7},
  bibsource    = {dblp computer science bibliography, https://dblp.org}
}

@inproceedings{DBLP:conf/bpm/KnochPS20,
  author       = {S{\"{o}}nke Knoch and
                  Shreeraman Ponpathirkoottam and
                  Tim Schwartz},
  editor       = {Dirk Fahland and
                  Chiara Ghidini and
                  J{\"{o}}rg Becker and
                  Marlon Dumas},
  title        = {Video-to-Model: Unsupervised Trace Extraction from Videos for Process
                  Discovery and Conformance Checking in Manual Assembly},
  booktitle    = {Business Process Management - 18th International Conference, {BPM}
                  2020, Seville, Spain, September 13-18, 2020, Proceedings},
  series       = {Lecture Notes in Computer Science},
  volume       = {12168},
  pages        = {291--308},
  publisher    = {Springer},
  year         = {2020},
  doi          = {10.1007/978-3-030-58666-9\_17},
  bibsource    = {dblp computer science bibliography, https://dblp.org}
}

@article{CHVIROVA2024110716,
title = {A multimedia dataset for object-centric business process mining in IT asset management},
journal = {Data in Brief},
volume = {55},
pages = {110716},
year = {2024},
issn = {2352-3409},
doi = {https://doi.org/10.1016/j.dib.2024.110716},
author = {Diana Chvirova and Andreas Egger and Tobias Fehrer and Wolfgang Kratsch and Maximilian Röglinger and Jakob Wittmann and Niklas Wördehoff}
}

@book{DBLP:books/sp/Weske19,
  author       = {Mathias Weske},
  title        = {Business Process Management - Concepts, Languages, Architectures,
                  Third Edition},
  publisher    = {Springer},
  year         = {2019},
  doi          = {10.1007/978-3-662-59432-2},
  isbn         = {978-3-662-59431-5},
  bibsource    = {dblp computer science bibliography, https://dblp.org}
}

@article{DBLP:journals/is/Scheer94,
  author       = {August{-}Wilhelm Scheer},
  title        = {{ARIS} Toolset: {A} Software Product is Born},
  journal      = {Inf. Syst.},
  volume       = {19},
  number       = {8},
  pages        = {607--624},
  year         = {1994},
  doi          = {10.1016/0306-4379(94)90031-0},
  bibsource    = {dblp computer science bibliography, https://dblp.org}
}

@inproceedings{DBLP:conf/ejc/WalterTAFL14,
  author       = {J{\"{u}}rgen Walter and
                  Tom Thaler and
                  Peyman Ardalani and
                  Peter Fettke and
                  Peter Loos},
  editor       = {Bernhard Thalheim and
                  Hannu Jaakkola and
                  Yasushi Kiyoki and
                  Naofumi Yoshida},
  title        = {Development and usage of a process model corpus},
  booktitle    = {Information Modelling and Knowledge Bases XXVI, 24th International
                  Conference on Information Modelling and Knowledge Bases {(EJC} 2014),
                  Kiel, Germany, June 3-6, 2014},
  series       = {Frontiers in Artificial Intelligence and Applications},
  volume       = {272},
  pages        = {437--448},
  publisher    = {{IOS} Press},
  year         = {2014},
  doi          = {10.3233/978-1-61499-472-5-437},
  bibsource    = {dblp computer science bibliography, https://dblp.org}
}

@inproceedings{DBLP:conf/icpm/SolaWSBRK22,
  author       = {Diana Sola and
                  Christian Warmuth and
                  Bernhard Sch{\"{a}}fer and
                  Peyman Badakhshan and
                  Jana{-}Rebecca Rehse and
                  Timotheus Kampik},
  editor       = {Marco Montali and
                  Arik Senderovich and
                  Matthias Weidlich},
  title        = {{SAP} Signavio Academic Models: {A} Large Process Model Dataset},
  booktitle    = {Process Mining Workshops - {ICPM} 2022 International Workshops, Bozen-Bolzano,
                  Italy, October 23-28, 2022, Revised Selected Papers},
  series       = {Lecture Notes in Business Information Processing},
  volume       = {468},
  pages        = {453--465},
  publisher    = {Springer},
  year         = {2022},
  doi          = {10.1007/978-3-031-27815-0\_33},
  bibsource    = {dblp computer science bibliography, https://dblp.org}
}

@inproceedings{DBLP:conf/bpm/BellanADGP22,
  author       = {Patrizio Bellan and
                  Han van der Aa and
                  Mauro Dragoni and
                  Chiara Ghidini and
                  Simone Paolo Ponzetto},
  editor       = {Cristina Cabanillas and
                  Niels Frederik Garmann{-}Johnsen and
                  Agnes Koschmider},
  title        = {{PET:} An Annotated Dataset for Process Extraction from Natural Language
                  Text Tasks},
  booktitle    = {Business Process Management Workshops - {BPM} 2022 International Workshops,
                  M{\"{u}}nster, Germany, September 11-16, 2022, Revised Selected
                  Papers},
  series       = {Lecture Notes in Business Information Processing},
  volume       = {460},
  pages        = {315--321},
  publisher    = {Springer},
  year         = {2022},
  doi          = {10.1007/978-3-031-25383-6\_23},
  bibsource    = {dblp computer science bibliography, https://dblp.org}
}

@article{bommasani2022opportunitiesrisksfoundationmodels,
      title={On the Opportunities and Risks of Foundation Models}, 
      author={Rishi Bommasani and Drew A. Hudson and Ehsan Adeli and Russ Altman and Simran Arora and Sydney von Arx and Michael S. Bernstein and Jeannette Bohg and Antoine Bosselut and Emma Brunskill and Erik Brynjolfsson and Shyamal Buch and Dallas Card and Rodrigo Castellon and Niladri Chatterji and Annie Chen and Kathleen Creel and Jared Quincy Davis and Dora Demszky and Chris Donahue and Moussa Doumbouya and Esin Durmus and Stefano Ermon and John Etchemendy and Kawin Ethayarajh and Li Fei-Fei and Chelsea Finn and Trevor Gale and Lauren Gillespie and Karan Goel and Noah Goodman and Shelby Grossman and Neel Guha and Tatsunori Hashimoto and Peter Henderson and John Hewitt and Daniel E. Ho and Jenny Hong and Kyle Hsu and Jing Huang and Thomas Icard and Saahil Jain and Dan Jurafsky and Pratyusha Kalluri and Siddharth Karamcheti and Geoff Keeling and Fereshte Khani and Omar Khattab and Pang Wei Koh and Mark Krass and Ranjay Krishna and Rohith Kuditipudi and Ananya Kumar and Faisal Ladhak and Mina Lee and Tony Lee and Jure Leskovec and Isabelle Levent and Xiang Lisa Li and Xuechen Li and Tengyu Ma and Ali Malik and Christopher D. Manning and Suvir Mirchandani and Eric Mitchell and Zanele Munyikwa and Suraj Nair and Avanika Narayan and Deepak Narayanan and Ben Newman and Allen Nie and Juan Carlos Niebles and Hamed Nilforoshan and Julian Nyarko and Giray Ogut and Laurel Orr and Isabel Papadimitriou and Joon Sung Park and Chris Piech and Eva Portelance and Christopher Potts and Aditi Raghunathan and Rob Reich and Hongyu Ren and Frieda Rong and Yusuf Roohani and Camilo Ruiz and Jack Ryan and Christopher Ré and Dorsa Sadigh and Shiori Sagawa and Keshav Santhanam and Andy Shih and Krishnan Srinivasan and Alex Tamkin and Rohan Taori and Armin W. Thomas and Florian Tramèr and Rose E. Wang and William Wang and Bohan Wu and Jiajun Wu and Yuhuai Wu and Sang Michael Xie and Michihiro Yasunaga and Jiaxuan You and Matei Zaharia and Michael Zhang and Tianyi Zhang and Xikun Zhang and Yuhui Zhang and Lucia Zheng and Kaitlyn Zhou and Percy Liang},
      year={2022},
      journal   = {arXiv preprint arXiv:2108.07258},
      archivePrefix={arXiv},
      primaryClass={cs.LG},
      }

@inproceedings{DBLP:conf/mkwi/ThalerHFL14,
  author       = {Tom Thaler and
                  Philip Hake and
                  Peter Fettke and
                  Peter Loos},
  editor       = {Dennis Kundisch and
                  Leena Suhl and
                  Lars Beckmann},
  title        = {Evaluating the Evaluation of Process Matching Techniques},
  booktitle    = {Multikonferenz Wirtschaftsinformatik, {MKWI} 2014, Paderborn, Germany, February 26-28, 2014},
  pages        = {1600--1612},
  publisher    = {University of Paderborn},
  year         = {2014},
  bibsource    = {dblp computer science bibliography, https://dblp.org}
}

@article{marx2022method,
  author    = {Marx, Vivien},
  title     = {Method of the Year: protein structure prediction},
  journal   = {Nature Methods},
  volume    = {19},
  number    = {2},
  pages     = {5--10},
  year      = {2022}
}

@inproceedings{thaler2016automated,
  author    = {Tom Thaler and Constantin Houy and Peter Fettke and Peter Loos},
  title     = {Automated Assessment of Process Modeling Exams: Basic Ideas and Prototypical Implementation},
  booktitle = {Modellierung 2016 - Workshopband. Workshop zur Modellierung in der Hochschullehre (MoHoL-2016)},
  editor    = {Stefanie Betz and Ulrich Reimer},
  series    = {Lecture Notes in Informatics (LNI)},
  volume    = {255},
  pages     = {63--70},
  publisher = {Gesellschaft für Informatik (GI)},
  address   = {Bonn},
  month     = {March},
  year      = {2016},
  isbn      = {978-3-88579-649-7}
}

@article{ullrich2023automated,
  author    = {Meike Ullrich and Constantin Houy and Tobias Stottrop and Michael Striewe and Brian Willems and Peter Fettke and Peter Loos and Andreas Oberweis},
  title     = {Automated Assessment of Conceptual Models in Education - A Systematic Literature Review},
  journal   = {Enterprise Modelling and Information Systems Architectures - International Journal of Conceptual Modeling (EMISAJ)},
  volume    = {18},
  number    = {2},
  pages     = {1--36},
  year      = {2023},
  publisher = {Gesellschaft für Informatik (GI)}
}

@article{kapoor2024ai,
  author    = {Sayash Kapoor and Benedikt Stroebl and Zachary S. Siegel and Nitya Nadgir and Arvind Narayanan},
  title     = {AI Agents That Matter},
  year      = {2024},
  journal   = {arXiv preprint arXiv:2401.00001},
  }

@article{kourani2024evaluating,
  author    = {Humam Kourani and Alessandro Berti and Daniel Schuster and Wil M. P. van der Aalst},
  title     = {Evaluating Large Language Models on Business Process Modeling: Framework, Benchmark, and Self-Improvement Analysis},
  year      = {2024},
  journal   = {arXiv preprint arXiv:2402.00001},
  }

@inproceedings{Westergaard2013180,
  author    = {Westergaard, M. and Fahland, D. and Stahl, C.},
  booktitle = {Proc. of 33rd International Conference on Application and Theory of Petri Nets and Other Models of Concurrency (Petri Nets 2012)},
  year      = {2013},
  doi       = {10.1007/978-3-642-40465-8_10},
  isbn      = {9783642404641},
  pages     = {180--202},
  title     = {{Grade/CPN: A tool and temporal logic for testing colored Petri net models in teaching}},
  volume    = {8100},
}

@article{Sanchez-Ferreres2020552,
  author    = {Sanchez-Ferreres, J. and Delicado, L. and Andaloussi, A. A. and Burattin, A. and Calderon-Ruiz, G. and Weber, B. and Carmona, J. and Padro, L.},
  publisher = {IEEE},
  year      = {2020},
  doi       = {10.1109/TLT.2020.2983916},
  issn      = {1939-1382},
  journal = {IEEE Transactions on Learning Technologies},
  number    = {3},
  pages     = {552--566},
  title     = {{Supporting the Process of Learning and Teaching Process Models}},
  volume    = {13},
}

@inproceedings{Schramm2012472,
  author    = {Schramm, J. and Strickroth, S. and Le, N.-T. and Pinkwart, N.},
  booktitle = {Proc. of 25th International Florida Artificial Intelligence Research Society Conference (FLAIRS-25)},
  year      = {2012},
  isbn      = {9781577355588},
  pages     = {472--477},
  title     = {{Teaching UML skills to novice programmers using a sample solution based intelligent tutoring system}},
}

@inproceedings{Striewe2014336,
  author    = {Striewe, M. and Goedicke, M.},
  publisher = {ACM},
  booktitle = {Proc. of 19th Annual Innovation and Technology in Computer Science Education Conference (ITiCSE 2014)},
  year      = {2014},
  doi       = {10.1145/2591708.2602657},
  isbn      = {9781450328333},
  pages     = {336},
  title     = {{Automated assessment of UML activity diagrams}},
}

@inproceedings{Beck2015,
  author    = {Beck, P.-D. and Mahlmeister, T. and Ifland, M. and Puppe, F.},
  booktitle = {Proc. of 2nd Workshop "Automatische Bewertung von Programmieraufgaben" (ABP 2015)},
  year      = {2015},
  series    = {CEUR-WS},
  title     = {{COCLAC - Feedback generation for combined UML class and activity diagram modeling tasks}},
  volume    = {1496},
}

@article{Fill2023Conceptual,
  title     = {Conceptual Modeling and Large Language Models: Impressions From First Experiments With ChatGPT},
  author    = {Fill, Hans-Georg and Fettke, Peter and K{\"o}pke, Julius},
  journal   = {Enterprise Modelling and Information Systems Architectures (EMISAJ) -- International Journal of Conceptual Modeling},
  volume    = {18},
  pages     = {1--15},
  year      = {2023},
  doi       = {10.18417/emisa.18.3},
  }

@article{DIJKMAN2011498,
title = {Similarity of business process models: Metrics and evaluation},
journal = {Information Systems},
volume = {36},
number = {2},
pages = {498-516},
year = {2011},
note = {Special Issue: Semantic Integration of Data, Multimedia, and Services},
issn = {0306-4379},
doi = {https://doi.org/10.1016/j.is.2010.09.006},
author = {Remco Dijkman and Marlon Dumas and Boudewijn {van Dongen} and Reina Käärik and Jan Mendling},
}

@article{bunke1998graph,
  author = {Bunke, Horst and Shearer, Kim},
  journal = {Pattern recognition letters},
  number = {3-4},
  pages = {255--259},
  publisher = {Elsevier},
  title = {A graph distance metric based on the maximal common subgraph},
  volume = 19,
  year = 1998
}

@article{norouzifar2024bridging,
  title     = {Bridging Domain Knowledge and Process Discovery Using Large Language Models},
  author    = {Norouzifar, Ali and Kourani, Humam and Dees, Marcus and van der Aalst, Wil M.P.},
  journal   = {arXiv preprint arXiv:2408.17316},
  year      = {2024},
  }

@incollection{VONROSING2015433,
title = {Business Process Model and Notation—BPMN},
editor = {Mark {von Rosing} and August-Wilhelm Scheer and Henrik {von Scheel}},
booktitle = {The Complete Business Process Handbook},
publisher = {Morgan Kaufmann},
address = {Boston},
pages = {433-457},
year = {2015},
isbn = {978-0-12-799959-3},
doi = {https://doi.org/10.1016/B978-0-12-799959-3.00021-5},
author = {Mark {von Rosing} and Stephen White and Fred Cummins and Henk {de Man}},
}

@article{kourani2024leveraginglargelanguagemodels,
      title={Leveraging Large Language Models for Enhanced Process Model Comprehension}, 
      author={Humam Kourani and Alessandro Berti and Jasmin Hennrich and Wolfgang Kratsch and Robin Weidlich and Chiao-Yun Li and Ahmad Arslan and Daniel Schuster and Wil M. P. van der Aalst},
      year={2024},
      journal   = {arXiv preprint arXiv:2408.08892},
      }

@InProceedings{10.1007/978-3-031-61007-3_18,
author="Kourani, Humam
and Berti, Alessandro
and Schuster, Daniel
and van der Aalst, Wil M. P.",
editor="van der Aa, Han
and Bork, Dominik
and Schmidt, Rainer
and Sturm, Arnon",
title="Process Modeling with Large Language Models",
booktitle="Enterprise, Business-Process and Information Systems Modeling. LNBIP 511",
year="2024",
publisher="Springer Nature Switzerland",
address="Cham",
pages="229-244",
}

@book{books/sp/Krogstie16,
  author = {Krogstie, John},
  ee = {http://dx.doi.org/10.1007/978-3-319-42512-2},
  isbn = {978-3-319-42512-2},
  pages = {1-250},
  publisher = {Springer},
  title = {Quality in Business Process Modeling},
  year = 2016
}

@InProceedings{10.1007/978-3-319-19237-6_25,
author="Heggset, Merethe
and Krogstie, John
and Wesenberg, Harald",
editor="Gaaloul, Khaled
and Schmidt, Rainer
and Nurcan, Selmin
and Guerreiro, S{\'e}rgio
and Ma, Qin",
title="Understanding Model Quality Concerns When Using Process Models in an Industrial Company",
booktitle="Enterprise, Business-Process and Information Systems Modeling",
year="2015",
publisher="Springer International Publishing",
address="Cham",
pages="395-409",
isbn="978-3-319-19237-6"
}

@InProceedings{10.1007/978-3-030-35646-0_11,
author="Pavlicek, Josef
and Pavlickova, Petra
and Naplava, Pavel",
editor="Pergl, Robert
and Babkin, Eduard
and Lock, Russell
and Malyzhenkov, Pavel
and Merunka, Vojt{\v{e}}ch",
title="Measures of Quality in Business Process Modeling",
booktitle="Enterprise and Organizational Modeling and Simulation",
year="2019",
publisher="Springer International Publishing",
address="Cham",
pages="146--155",
isbn="978-3-030-35646-0"
}

@InProceedings{10.1007/978-3-031-77908-4_7,
author="Gutschmidt, Anne
and Nast, Benjamin",
editor="Paja, Elda
and Zdravkovic, Jelena
and Kavakli, Evangelia
and Stirna, Janis",
title="Assessing Model Quality Using Large Language Models",
booktitle="The Practice of Enterprise Modeling",
year="2025",
publisher="Springer Nature Switzerland",
address="Cham",
pages="105--122",
isbn="978-3-031-77908-4"
}

@InProceedings{10.1007/978-3-642-34002-4_5,
author="Houy, Constantin
and Fettke, Peter
and Loos, Peter",
editor="Atzeni, Paolo
and Cheung, David
and Ram, Sudha",
title="Understanding Understandability of Conceptual Models -- What Are We Actually Talking about?",
booktitle="Conceptual Modeling",
year="2012",
publisher="Springer Berlin Heidelberg",
address="Berlin, Heidelberg",
pages="64--77",
isbn="978-3-642-34002-4"
}

@misc{joel2024surveyllmbasedcodegeneration,
      title={A Survey on LLM-based Code Generation for Low-Resource and Domain-Specific Programming Languages}, 
      author={Sathvik Joel and Jie JW Wu and Fatemeh H. Fard},
      year={2024},
      journal   = {arXiv preprint arXiv:2410.03981}, 
}

@misc{sarkar2022likeprogramartificialintelligence,
      title={What is it like to program with artificial intelligence?}, 
      author={Advait Sarkar and Andrew D. Gordon and Carina Negreanu and Christian Poelitz and Sruti Srinivasa Ragavan and Ben Zorn},
      year={2022},
      journal   = {arXiv preprint arXiv:2208.06213}, 
}

@article{katz2024thought,
  title={Thought of Search: Planning with Language Models Through The Lens of Efficiency},
  author={Katz, Michael and Kokel, Harsha and Srinivas, Kavitha and Sohrabi Araghi, Shirin},
  journal={Advances in Neural Information Processing Systems},
  volume={37},
  pages={138491-138568},
  year={2024}
}


\end{document}